\documentclass[conf]{new-aiaa}
\usepackage[utf8]{inputenc}

\usepackage{graphicx}
\usepackage{amsmath}
\usepackage[version=4]{mhchem}
\usepackage{siunitx}
\usepackage{longtable,tabularx}
\usepackage{subfigure}
\usepackage{float}

\usepackage{bm}
\usepackage{bbm}

\newcommand*{\QED}{\hfill\ensuremath{\square}}%

\renewcommand{\vec}[1]{\bm{#1}}
\newtheorem{theorem}{Proposition}

\title{Reduced-order modeling using Dynamic Mode Decomposition and Least Angle Regression}

\author{John Graff,\footnote{Graduate student, AIAA Student Member} Xianzhang Xu,\footnote{Graduate student, AIAA Student Member} Francis D. Lagor,\footnote{Assistant Professor, AIAA Member, email: flagor@buffalo.edu} and Tarunraj Singh\footnote{Professor, AIAA Associate Fellow, email: tsingh@buffalo.edu}}
\affil{Department of Mechanical and Aerospace Engineering,\\ University at Buffalo, The State University of New York, Buffalo, NY, 14260}

\begin{document}

\begin{minipage}[t]{\textwidth}
\vspace{-.5in}
Citation: J. Graff, X. Xu, F. D. Lagor, and T. Singh. Reduced-order modeling using Dynamic Mode Decomposition and Least Angle Regression. AIAA Aviation Forum, 3072, 2019.\\

Source code available at:  http://driftlab.eng.buffalo.edu/code.html
\end{minipage}

\maketitle

\begin{abstract}
Dynamic Mode Decomposition (DMD) yields a linear, approximate model of a system's dynamics that is built from data. We seek to reduce the order of this model by identifying a reduced set of modes that best fit the output. We adopt a model selection algorithm from statistics and machine learning known as Least Angle Regression (LARS). We modify LARS to be complex-valued and utilize LARS to select DMD modes. We refer to the resulting algorithm as Least Angle Regression for Dynamic Mode Decomposition (LARS4DMD). Sparsity-Promoting Dynamic Mode Decomposition (DMDSP), a popular mode-selection algorithm, serves as a benchmark for comparison. Numerical results from a Poiseuille flow test problem show that LARS4DMD yields reduced-order models that have comparable performance to DMDSP. LARS4DMD has the added benefit that the regularization weighting parameter required for DMDSP is not needed.
\end{abstract}

%%%%%%%%%%%%%%%%%%%%%%%%%%%%%%%%%%%%%%%%%%%%%%%%%%%%%%%%%%%%%%%%%%%%%%%%%%%%%%%%%%%%
\section{Introduction}\label{sec:intro}

\lettrine{D}{ata}-driven system analysis has become an increasingly popular technique for studying features of dynamical systems. Dynamic Mode Decomposition (DMD) is a data analysis algorithm that can identify dynamical features that appear in the output data of a system \cite{Schmid2010}. DMD works by finding the best-fit linear operator $A$ that marches the measurements forward in time. The best-fit linear operator can be analyzed through modal analysis and its eigenvectors are known as the DMD modes of the system with corresponding DMD eigenvalues. These modes can be utilized to give insight into features of dynamical systems such as coherent structures in fluid flows \cite{ZHANG_2014} and patterns in neural recordings \cite{BRUNTON_2016}. It has also been shown that there is a strong connection between DMD and Koopman operator theory \cite{koopman}.

DMD first appeared in the fluids literature \cite{Schmid2010} and is particularly useful for data-driven analysis of fluid systems \cite{Brunton_AIAA_2019, LIU_2019, Pan_2017}. For fluid systems, DMD is frequently used to analyze Particle Image Velocimetry (PIV) data, which is a measurement technique that creates vector-field images of a flowfield in time. Measurement techniques such as PIV often require very fine spatial resolution to resolve the fluid dynamics \cite{Tu2014}. When the number of spatial data points is much larger than the total number of temporal observations or snapshots, the number of snapshots often determines the number of DMD modes. Including more snapshots can help to capture the system's dynamics in the DMD modes. However, since the number of modes increases with the number of snapshots, often the DMD model is large. To deal with a large number of DMD modes, one typically reduces the size of the model by retaining only relatively important modes. Determining the DMD modes that are most important is a reduced-order modeling problem and is the focus of this paper.

A reduced-order modeling algorithm that has been effective in creating sparse DMD models is Sparsity-Promoting Dynamic Mode Decomposition (DMDSP) \cite{DMDSP, Emotion, Phonon}. DMDSP has also been extended to systems with inputs \cite{DMDSP_inputs}. It has been used to cancel DMD modes that are associated with noise \cite{Dawson2016} and for model selection in filtering applications \cite{Gomez}. The authors of \cite{DMDSP} observe that the reduced-order modeling problem amounts to appropriately weighting DMD modes in the model, because a mode's contribution can be neglected by scaling its weight to zero. One process for weighting modes involves optimizing the reconstruction of the original data using the model. The reconstruction problem seeks to minimize the reconstruction error, which is the difference between the full data set and the data predicted by the model. DMDSP creates a reduced-order model through the multi-objective optimization of minimizing the reconstruction error and the $l_1$ norm of the vector of mode amplitudes. The $l_1$ regularization encourages sparsity in the mode amplitude vector by scaling the mode amplitudes to zero for modes that have minimal influence on the reconstruction error. The final step in DMDSP removes the regularization term and re-solves for the mode amplitudes for the selected set of modes.

The number of modes that DMDSP deactivates depends on the relative weighting between the reconstruction error and the $l_1$ regularization term in the optimization. The relative weighting is mediated by a user-specified weighting coefficient. Unfortunately, the user does not know values of this parameter a priori that will produce useful model reconstructions. Further, the user does not know how the weighting parameter will influence the induced sparsity and the associated tradeoff in performance. The user therefore searches for a range of regularization values for which DMDSP provides model reduction. The user then examines the resulting models and selects the one that produces an acceptable tradeoff between performance and model size.

In this work, we propose a new method for determining the DMD modes that are most relevant to the system dynamics through adaption of the Least Angle Regression (LARS) algorithm \cite{LARS}. LARS is a regression algorithm that sequentially selects vectors from a set of candidate vectors and appropriately scales them to fit a vector of data. These user-specified candidate vectors for model construction are called covariates. As an example, if the measurement data is a disease diagnosis in a set of patients, the covariates may consist of other relevant data collected about the patients such as their age, height, weight, and so forth. LARS determines which of these covariates are important and scales the selected covariates appropriately to yield an estimate of the diagnosis.

LARS is a member of a class of algorithms known as forward selection algorithms that build predictive models in a stepwise manner. The "S" in the LARS acronym references the popular Stagewise and Lasso algorithms that are shown to be variants of LARS \cite{LARS}. These algorithms sequentially build up a model by traveling in the space of covariates along directions that are determined by the covariates that are most correlated with the residual (i.e., the difference between the data and the current model's prediction of the data) \cite{LARS}. The direction of travel that LARS selects is the equiangular direction between the most correlated covariates \cite{LARS}. We choose to apply a forward selection algorithm for DMD mode selection to create a principled approach for examining the tradeoff between performance and model size during reduced-order model construction. Using a forward selection algorithm, a user can more easily track the influence of an individual mode on the model.

LARS functions by sequentially adding covariates to an active set based on their correlation with the residual. After each selection step, LARS generates an interim estimate of the data that is formed by traveling along an equiangular direction with all covariates in the active set. The contribution of each covariate to travel along the equiangular direction determines its regression coefficient, which weights the covariate in the model.

This paper contributes a modified version of the LARS algorithm, known as Complex LARS, that handles complex-valued data and complex-valued covariates, and the LARS4DMD algorithm that applies Complex LARS for reduced-order DMD model construction. The LARS algorithm was originally developed for data fitting in statistics using real-valued data and real-valued covariates. We adapt LARS to operate using complex data and complex covariates by replacing the dot product with a complex-valued inner product over the vector space $\mathbbm{C}^n$. These contributions are significant because they enhance automation in constructing reduced-order DMD models by eliminating the need for a problem-dependent regularization weighting parameter. The performance of the LARS4DMD algorithm is demonstrated on synthesized Poiseuille flow data that is available from the original DMDSP paper \cite{DMDSP,DMDSP_site}.

The outline of this paper is as follows. Section \ref{sec:background} presents necessary background on the DMD, DMDSP, and LARS algorithms. Section \ref{sec:C_LARS} derives Complex LARS. Section \ref{sec:LARS_DMD} applies Complex LARS for use in DMD mode selection. Section \ref{sec:results} shows the effectiveness of LARS4DMD in generating reduced-order DMD models using DMDSP as a benchmark. Section \ref{sec:conclusion} concludes the paper and discusses ongoing work.

%%%%%%%%%%%%%%%%%%%%%%%%%%%%%%%%%%%%%%%%%%%%%%%%%%%%%%%%%%%%%%%%%%%%%%%%%%%%%%%%%%%
\section{Data-driven, reduced-order modeling} \label{sec:background}

This section presents the techniques necessary for development of LARS4DMD: Section \ref{subsec:DMD} reviews DMD; Section \ref{subsec:DMDSP} describes DMDSP, a state-of-the-art method for DMD mode selection; and Section \ref{subsec:LARS} introduces the original LARS algorithm.

\subsection{Dynamic Mode Decomposition (DMD)} \label{subsec:DMD}

The DMD data analysis begins with the collection and proper arrangement of measurements for processing. Although generalized definitions of DMD exist (e.g., see \cite{Tu2014, DMD_Book}), we focus on the case of sequential, constant-interval measurements of a process evolving in time, similar to \cite{DMDSP}. Let $\vec{\psi}_k$ be a measurement vector (or snapshot) of the system for time steps $k = 0,\dots,N$. DMD seeks the best-fit linear operator $A$ that advances each snapshot one time step such that $\vec{\psi}_{k+1} \approx A\vec{\psi}_{k}$. By constructing two data matrices, $\Psi_0 = [\begin{array}{cccc} \vec{\psi}_0 & \vec{\psi}_1 & \dots & \vec{\psi}_{(N-1)}\end{array}]$ and $\Psi_1 = [\begin{array}{cccc}\vec{\psi}_1 & \vec{\psi}_2 & \dots & \vec{\psi}_N\end{array}]$, that have column entries offset by one timestep, this condition can be expressed as $\Psi_1 \approx A\Psi_0$ \cite{DMDSP}. The DMD optimization problem \cite{DMDSP}
\begin{equation}
\min_A\quad || \Psi_1 - A\Psi_0||^2_{\cal F},
\label{eq:opt}
\end{equation}
where ${\cal F}$ is the Frobenius norm, provides the best-fit $A$ matrix $A^\ast = \Psi_1\Psi_0^\dagger$, where $(\cdot)^\dagger$ is the Moore-Penrose pseudo-inverse \cite{DMDSP}.

The eigenvectors of $A$ are the DMD modes with associated DMD eigenvalues. In practice, the $A$ matrix can often be too large to form \cite{Schmid2010}. When this occurs, the DMD modes and eigenvalues are still accessible by first solving for a projected version of $A$. Let $\Psi_0=U\Sigma V^H$ be an economy Singular Value Decomposition (SVD) of $\Psi_0$, where $(\cdot)^H$ denotes the Hermitian or conjugate-transpose operation, and let $r$ be the rank of $\Psi_0$. To address the possibility of linearly dependant snapshots in $\Psi_0$, truncate $U$, $\Sigma$, and $V$ according to $r$. Let $U_r$ be the first $r$ columns of $U$, let $\Sigma_r$ be an $r \times r$ matrix extracted from the upper left corner of $\Sigma$, and let $V_r$ be the first $r$ columns of $V$.

Consider a version $F$ of the $A$ matrix that is transformed into the basis formed by the columns of $U_r$ such that \cite{DMDSP}
\begin{equation}
A \approx U_r FU_r^H.
\label{eq:Ah}
\end{equation}
Inserting \eqref{eq:Ah} and the SVD of $\Psi_0$ into \eqref{eq:opt} and optimization yields the best-fit projected version of the A matrix, given by \cite{DMDSP}
\begin{equation}
F_\text{DMD} = U_r^H\Psi_1V_r\Sigma_r^{-1}. \label{eq:Fdmd}
\end{equation}
Performing modal analysis on $F_\text{DMD}$ can provide insight on the DMD modes of $A$. Let $\vec{v}_j$ be and eigenvector of $F_\text{DMD}$ such that $F_\text{DMD}\vec{v}_j = \lambda_j\vec{v}_j$ for eigenvalue $\lambda_j$. A DMD mode $\vec{\phi}_j$ of $A$ can be recovered from the corresponding eigenvector $\vec{v}_j$ in the subspace spanned by the columns of $U_r$ by \cite{DMDSP}
\begin{equation}
\vec{\phi}_j = U_r\vec{v}_j.\label{eq:DMDmode}
\end{equation}
Using the DMD modes, it is possible to approximate the dynamics of the measurements or snapshots. A measurement vector $\vec{\psi}_k$ at time $k$ can be expressed as a linear combination of the DMD modes. The DMD modes evolve in time by repeated multiplication with their corresponding DMD eigenvalues, yielding the snapshot dynamics \cite{DMDSP}
\begin{equation}
\vec{\psi}_k \approx \sum_{j = 1}^r\vec{\phi}_j\lambda_j^k\alpha_j,
\label{eq:rec}
\end{equation}
where $\alpha_j$ is, in general, a complex scalar that corresponds to the contribution of the mode $\vec{\phi}_j$ to the initial snapshot matrix $\Psi_0$. The amplitudes $\alpha_j$ have also been shown to be equivalent to the values of the Koopman eigenfunctions calculated at the initial condition \cite{koopman}.

Using the snapshot dynamics \eqref{eq:rec}, it is possible to reconstruct the data matrix $\Psi_0$ from an initial snapshot vector by letting $\vec{\psi}_k^\text{rec} = \vec{\psi}_k$ for each timestep $k$. The DMD mode amplitudes are computed using $\alpha_j =\vec{v}_j^H U_r^H \vec{\psi}_0$. Solving for the DMD mode amplitudes to best-fit DMD modes to a data set is referred to as the reconstruction problem. In the context of DMD, reduced-order modeling seeks to identify a subset of the DMD modes that perform well in data reconstruction for a dataset or a variety of datasets.

\subsection{Sparsity-Promoting Dynamic Mode Decomposition (DMDSP)} \label{subsec:DMDSP}

DMDSP is a reduced-order modeling technique that selects modes by increasing the sparsity in the vector of DMD amplitudes $\vec{\alpha}$ during a reconstruction optimization problem. To state the reconstruction problem mathematically, put the DMD modes $\vec{\phi}_j$ for $j=1,\dots,r$ into matrix form
\begin{align}
\Phi =
\left[\begin{array}{cccc}
\vec{\phi}_1&\vec{\phi}_2&\dots&\vec{\phi}_r
\end{array}\right],
\label{eq:phi}
\end{align}
Using the DMD eigenvalues $\lambda_j$, construct a Vandermonde matrix
\begin{align}
\Xi =
%\begin{bmatrix}
\left[
\begin{array}{cccc}
\lambda_1^{0}&\lambda_1^{1}&\dots&\lambda_1^{N-1}\\
\lambda_2^{0}&\lambda_2^{1}&\dots&\lambda_2^{N-1}\\
\vdots&\vdots&\ddots&\vdots \\
\lambda_r^{0}&\lambda_r^{1}&\dots&\lambda_r^{N-1}\\
\end{array}
\right]
%\end{bmatrix},
\label{eq:vand}
\end{align}
which expresses the temporal evolution of the eigenvalue portion of each mode's coefficient in the snapshot dynamics \eqref{eq:rec}. Form a diagonal matrix that is constructed from the DMD mode amplitudes such that $D_\alpha = \text{diag}([\begin{array}{cccc} \alpha_1 & \alpha_2 & \dots & \alpha_r \end{array}])$. Using the diagonal amplitude matrix $D_\alpha$, the DMD mode matrix \eqref{eq:phi}, and the Vandermonde matrix \eqref{eq:vand}, we can reconstruct $\Psi_0$ using the snapshot dynamics \cite{DMDSP}
\begin{equation}
\Psi_{0}^{\text{rec}} = \Phi\;D_\alpha\;\Xi.
\label{eq:DMD_rec}
\end{equation}
The reconstruction problem can be expressed as $$\min_{\vec{\alpha}} \quad J(\vec{\alpha}) = ||\Psi_0 - \Phi D_\alpha \Xi||^2_{\cal F}.$$

DMDSP is a two-step procedure that minimizes $ J(\vec{\alpha}) + \beta||\vec{\alpha}||_1$, where $||\vec{\alpha}||_1$ is an $l_1$ regularization penalty that serves as a proxy for penalizing the number of nonzero entries in $\vec{\alpha}$, and $\beta$ is a user-defined, regularization weighting term. The second step in DMDSP is a polishing step in which the algorithm re-solves for the mode amplitudes $\vec{\alpha}$ that best solve the reconstruction problem, but the regularization term is not present and the desired sparsity structure from the first step is strictly enforced. Typically, a user of DMDSP considers an array of regularization weights $\beta$. For each $\beta$ value, DMDSP solves for a mode amplitude vector $\vec{\alpha}$ that contains zeros for modes deactivated by the method and nonzero values for the amplitudes of selected modes. Each $\beta$ value therefore corresponds to a separate reduced-order model. However, it should be noted that it is possible, and common in practice, multiple $\beta$ values to yield the same reduced-order model.

The authors of \cite{DMDSP} define the percent performance loss to be a measure of the error in reconstruction, normalized by size of the original data set such that $$P_{\text{loss}} = 100\times\frac{||\Psi_0 - \Psi_0^\text{rec}||_{\cal F}}{||\Psi_0||_{\cal F}}.$$ The user can calculate the performance loss for each model, examine the performance and model size tradeoff, and select a model appropriate for the application.

\subsection{Least Angle Regression (LARS)} \label{subsec:LARS}

\indent Let $X = [\vec{x}_1\;\vec{x}_2\;\dots\;\vec{x}_r]$ be a set of zero-mean, unit-variance, and linearly independent covariates $\vec{x}_j$ for $j = 1,\;\dots,\;r$, and let $\vec{y}$ represent a zero-mean data vector \cite{LARS}. The LARS algorithm searches for regression coefficients $\alpha_j$ for $j = 1,\dots,r$, to construct a linear estimate
\begin{equation}
\vec{\mu} = \sum_{j=1}^r \vec{x}_j \alpha_j.
\label{eq:mu}
\end{equation}
Note that we use $\alpha_j$ notation to denote the DMD mode amplitudes and the LARS regression coefficients; these quantities correspond when we apply LARS4DMD for mode selection in Section \ref{sec:LARS_DMD}.

Algorithm 1 presents the sequential procedure for selection of covariates by LARS \cite{LARS}. In each iteration of the algorithm, LARS produces an estimate $\vec{\hat{\mu}}_S$ of the data based on currently selected covariates in a set of active covariates. The hat $\hat{(\cdot)}$ notation denotes the current iteration, and the subscript $(\cdot)_S$ indicates that the quantity is based on selected covariates in the active set. The current estimate initializes with $\vec{\hat{\mu}}_S = 0$. At each iteration, the difference between the data $\vec{y}$ and the current estimate $\vec{\hat{\mu}}_S$ is the residual $(\vec{y}-\vec{\hat{\mu}}_S)$. LARS calculates the current correlation between each covariate $\vec{x}_j$ and the residual $(\vec{y}-\vec{\hat{\mu}}_S)$ in Step 1.1. After finding the maximum absolute current correlation $\hat{C} = \max_j|\hat{c}_j|$ in Step 1.2, LARS selects covariates for which $\hat{C} = |\hat{c}_j|$ (or covariates for which this equality holds within a small tolerance). When adding covariate $\vec{x}_j$ to the active set, LARS multiplies each selected covariate by the sign of its correlation with the residual $\text{sgn}(\hat{c}_j)$ so that all covariates in the active set have positive correlations with the residual.\\

\begin{center}
\textbf{Algorithm 1 (LARS)}
\end{center}
\noindent Inputs: Zero-mean, real data vector $\vec{y}$, and a set of real, zero-mean, unit-variance covariates $X$.
\begin{enumerate}

\item[1.1)] Obtain a vector of correlations with the current residual $(\vec{y}-\vec{\hat{\mu}}_S)$, $$\vec{\hat{c}} = X^T(\vec{y}-\vec{\hat{\mu}}_S).$$

\item[1.2)] Find the current maximum absolute correlation, $\hat{C} = \max_j\left|\hat{c}_j\right|$.

\item[1.3)] Form the active set $S = \{ j \in 1,...,r \,\,\lvert\,\, |\hat{c}_j| = \hat{C}\}$, and collect the signum-aligned covariates $s_j\vec{x}_j$ of the active set, where $s_j = \text{sgn}(\hat{c}_j)$, within the columns of the signum-aligned covariate matrix $X_S = [\begin{array}{ccc} \dots & s_j\vec{x}_j &\dots \end{array}]$.

\item[1.4)] Create an equiangular direction of travel $\vec{u}_S$,
\begin{align*}
L_S &= \left(\vec{\mathbbm{1}}^T\left(X_S^T X_S\right)^{-1}\vec{\mathbbm{1}}\right)^{-1/2}\!\!,\\
\vec{w}_S &= \left(X_S^T X_S\right)^{-1}L_S\vec{\mathbbm{1}},\\
\vec{u}_S &= X_S \vec{w}_S.
\end{align*}

\item[1.5)] Find correlations with the direction of travel for all covariates
$$\vec{g} = X^T\vec{u}_S.$$

\item[1.6)] Find the length $\hat{\gamma}$ to travel along $\vec{u}_S$
\[
\hat{\gamma} = \left\{ \begin{array}{ll}\min^+_{j \in S^c} \quad \left\{ \frac{\hat{C}-\hat{c}_j}{L_S - g_j}, \frac{\hat{C}+\hat{c}_j}{L_S + g_j} \right\} & \text{if $S^c$ is nonempty,}\\
\frac{\tilde{C}}{L_S} & \text{if $S^c$ is empty,}
\end{array}\right.
\]
where $S^c$ is the complement of $S$ and $\min^+$ indicates the minimum taken over positive values only.

\item[1.7)] Update the estimate of the data,
$$\vec{\hat{\mu}}_{S,k} = \vec{\hat{\mu}}_{S,k-1} + \hat{\gamma}\vec{u}_S.$$
\emph{(Note: Initialize with $\vec{\hat{\mu}}_{S,0} = 0$.)}

\item[1.8)] Update the regression coefficients for $j = 1,\dots,r$,
\[
\alpha_{k,j} = \left\{ \begin{array}{ll}\alpha_{k-1,j} + \hat{\gamma} s_j w_{S,j}  &\text{if } j\in S,\\
0 &\text{if } j\notin S
\end{array}\right.
\]
\emph{(Note: Initialize with $\vec{\alpha}_{0} = 0$.)}

%\item[1.8.b)] Optional: (LARS-OLS hybrid method) Compute the regression coefficients using least squares, $$\vec{\alpha}_{\text{OLS},k} = \tilde{X}_S^\dagger \vec{y},$$ where $\tilde{X}_S$ is the matrix of selected, non-signum-aligned covariates.

\item[1.9)] Repeat Steps 1.1-1.8 until all covariates have zero correlation with the residual or until there are no covariates remaining in $S^c$.

\noindent Output: Vector of regression coefficients $\vec{\alpha}$.\\
\bigskip
\end{enumerate}

Using the covariates in the active set, LARS determines an equiangular direction of travel $\vec{u}_S$ in which LARS can step to reduce the current residual. To calculate the step direction, the equiangular condition \cite{LARS}
\begin{equation}
X_S^T \vec{u}_S = L_S \vec{\mathbbm{1}},
\label{eq:equi-angle}
\end{equation}
where $\vec{\mathbbm{1}}$ is a vector of ones, ensures that the dot products between each aligned covariate in the active set $s_j \vec{x}_j$ and the direction $\vec{u}_S$ have an equal value $L_S$. Requiring $\vec{u}_S$ to be a unit vector, i.e. $\vec{u}_S^T \vec{u}_S = 1$, one can utilize \eqref{eq:equi-angle} to derive the value \cite{LARS}
\begin{equation}
L_S = \left(\vec{\mathbbm{1}}^T\left(X_S^T X_S\right)^{-1}\vec{\mathbbm{1}}\right)^{-1/2},
\label{eq:L}
\end{equation}
in the equi-angle condition \eqref{eq:equi-angle}. $L_S$ used to calculate a vector of weighting coefficients $\vec{w}_S$ for the active covariates and the associated equi-angle direction such that  \cite{LARS}
\begin{equation}
\vec{w}_S = \left(X_S^T X_S\right)^{-1}L_S\vec{\mathbbm{1}},
\label{eq:w}
\end{equation}
and
\begin{equation*}
\vec{u}_S = X_S \vec{w}_S.
\end{equation*}
Traveling along the equiangular direction reduces the current correlation equally among all covariates in the active set \cite{LARS}. LARS selects a step size to travel in the equi-angle direction $\vec{u}_S$ that is as small as possible until another covariate enters the active set. Step 1.6 chooses the distance $\hat{\gamma}$ to travel along the equiangular direction $\vec{u}_S$. Subsequently, Step 1.7 uses $\hat{\gamma}\vec{u}_S$ to update the current estimate, and Step 1.8 provides the new regression coefficients. %Step 1.8.a offers the LARS method for determining the regression coefficients. Step 1.8.b provides another method known as the LARS-OLS hybrid method \cite{LARS}. The LARS-OLS hybrid method takes the covariates selected to be in the model by LARS and re-fits their regression coefficients using OLS.
The LARS algorithm repeats until all covariates have zero correlation with the residual or until all candidate covariates have joined the active set.  The algorithm returns the regression coefficients needed for model construction.  

%%%%%%%%%%%%%%%%%%%%%%%%%%%%%%%%%%%%%%%%%%%%%%%%%%%%%%%%%%%%%%%%%%%%%%%%%%%%%%%%

\section{Complex Least Angle Regression} \label{sec:C_LARS}

As formulated, LARS selects real covariates to fit real data. Often complex covariates can arise in applications, such as in the DMD mode-selection problem that this paper studies. Complex data can also occur, such as if the components of a planar vector field are stored together in complex format. The example in Section \ref{sec:results} uses complex data.

This section modifies LARS to allow for complex covariates and complex data. The adaption of LARS for complex covariates and complex data requires replacement of the inner product space over which the algorithm evolves. An inner product can be used to describe the angular relationship between two vectors, so its selection is important for the LARS algorithm, which seeks an equiangular direction of travel during covariate selection. Let covariates $\vec{x}_j$, for $j = 1,\dots,r$, and data vector $\vec{y}$ reside in the vector space $\mathbbm{C}^n$. Similar to LARS, let $\vec{y}$ be a zero-mean data vector and let $\vec{x}_j$ for $j=1,\dots,r$ be zero-mean, unit-variance\footnote{The variance for a vector of complex entries is $\text{var}(\vec{q}) = 1/n\sum_{j=1}^n|q_j-\text{mean}(\vec{q})|^2$.} covariates. Consider an inner product of the form \cite{inner_prod}
$$ \left<\cdot,\cdot\right> : \mathbbm{C}^n \times \mathbbm{C}^n \longrightarrow \mathbbm{C},$$
with the following properties for $\vec{x},\vec{y},\vec{z} \in \mathbbm{C}^n$ and $ a \in \mathbbm{C}$ \cite{inner_prod}:

\begin{enumerate}
\item[(i)] positive definiteness \\
\indent \hspace*{1cm} $\left<\vec{x},\vec{x}\right> \geq 0, \text{ with } \left<\vec{x},\vec{x}\right>=0 \text{ if and only if } \vec{x} = 0,$
\item[(ii)] conjugate symmetry \\
\indent \hspace*{1cm} $\left<\vec{x},\vec{y}\right> = \overline{\left<\vec{y},\vec{x}\right>},$
\item[(iii)] right linearity \\
\indent \hspace*{1cm} $\left<\vec{x},a\vec{y}\right> = a\left<\vec{x},\vec{y}\right>,$\\
\indent \hspace*{1cm} $\left<\vec{x},\vec{y}+\vec{z}\right> = \left<\vec{x},\vec{y}\right> + \left<\vec{x},\vec{z}\right>,$
\item[(iv)] left-conjugate linearity \\
\indent \hspace*{1cm} $\left<a\vec{x},\vec{y}\right> = \overline{a}\left<\vec{x},\vec{y}\right>,$\\
\indent \hspace*{1cm} $\left<\vec{x}+\vec{y},\vec{z}\right> = \left<\vec{x},\vec{z}\right> + \left<\vec{y},\vec{z}\right>.$
\end{enumerate}

Note that Property (iv) is a consequence of Properties (ii) and (iii). Left-conjugate linearity represents a choice of inner product convention that is often called the quantum mechanical definition; right-conjugate linearity is another common choice \cite{inner_prod}. We select the Euclidean inner product over $\mathbbm{C}^n$ defined by \cite{herm_prod}\\

\begin{equation}
\left<\vec{x},\vec{y}\right> =\vec{x}^H\vec{y},
\label{eq:IP}
\end{equation}
to satisfy left-conjugate linearity and Properties (i) through (iii).

Algorithm 2 presents Complex LARS. Each step of Complex LARS closely reflects the corresponding step of the LARS Algorithm. Step 1.1 in the LARS algorithm calculates a vector of correlations $\vec{\hat{c}}$ based on the dot product $\vec{x}_j^T \left(\vec{y}-\vec{\hat{\mu}}_S\right)$ between each covariate $\vec{x}_j$, for $j = 1,\dots,r$, and the current residual $(\vec{y}-\vec{\hat{\mu}}_S)$. Substitution of the inner product \eqref{eq:IP} leaves the first step unchanged except for replacing the transpose operation $(\cdot)^T$ with the Hermitian operation $(\cdot)^H$.  Note that the choice of inner product causes minimal change to the equations in Steps 2.1 and 2.4. Although the inner product \eqref{eq:IP} yields a complex value, we still refer to the elements of $\vec{\hat{c}}$ as correlations to retain the terminology from \cite{LARS}.

\begin{center}
\textbf{Algorithm 2 (Complex LARS)}\\
\end{center}
\noindent Inputs: Zero-mean data vector $\vec{y}$, and a set of zero-mean, unit-variance covariates $X$.
\begin{enumerate}

\item[2.1)] Obtain a vector of correlations with the current residual $(\vec{y}-\vec{\hat{\mu}}_S)$, $$\vec{\hat{c}} = X^H(\vec{y}-\vec{\hat{\mu}}_S).$$

\item[2.2)] Find the current maximum absolute correlation $\hat{C} = \max_j\left(|\hat{c}_j|\right)$.

\item[2.3)] Form the active set $S = \{ j \in 1,...,r \,\,\lvert\,\, |\hat{c}_j| = \hat{C}\}$, and collect the signum-aligned covariates $s_j\vec{x}_j$ of the active set, where $s_j = \text{sgn}_\mathbbm{C}(\hat{c}_j)$, within the columns of the signum-aligned covariate matrix $X_S = [\begin{array}{ccc} \dots & s_j\vec{x}_j &\dots \end{array}]$.

\item[2.4)] Create an equiangular direction of travel
\begin{align*}
L_S &= \left(\vec{\mathbbm{1}}^T\left(X_S^H X_S\right)^{-1}\vec{\mathbbm{1}}\right)^{-1/2},\\
\vec{w}_S &= \left(X_S^H X_S\right)^{-1}L_S\vec{\mathbbm{1}},\\
\vec{u}_S &= X_S \vec{w}_S.
\end{align*}
\item[2.5)] Find correlations with the direction of travel for all covariates
$$\vec{g} = X^H\vec{u}_S.$$

\item[2.6)] Find the length $\hat{\gamma}$ to travel along $\vec{u}_S$
\[
\hat{\gamma} = \left\{ \begin{array}{ll} \min^+_{j \in S^c} \quad \frac{\left(\Re\left\{\left<g_j,\hat{c}_j\right>\right\}-\hat{C}L_S\right) \pm \sqrt{\left(\Re\left\{\left<g_j,\hat{c}_j\right>\right\}-\hat{C}L_S\right)^2 - \left(|g_j|^2 - L_S^2\right)\left(|\hat{c}_j|^2 - \hat{C}^2\right)}}{|g_j|^2 - L_S^2}& \text{if $S^c$ is nonempty,}\\
\frac{\tilde{C}}{L_S} & \text{if $S^c$ is empty,}
\end{array}\right.
\]
where $S^c$ is the complement of $S$ and $\min^+$ indicates the minimum taken over positive values only.

\item[2.7)] Update the estimate of the data,
$$\vec{\hat{\mu}}_{S,k} = \vec{\hat{\mu}}_{S,k-1} + \hat{\gamma}\vec{u}_S.$$
\emph{(Note: Initialize with $\vec{\hat{\mu}}_{S,0} = 0$.)}

\item[2.8)] Update the regression coefficients for $j = 1,\dots,r$,
\[
\alpha_{k,j} = \left\{ \begin{array}{ll}\alpha_{k-1,j} + \hat{\gamma} s_j w_{S,j}  &\text{if } j\in S,\\
0 &\text{if } j\notin S
\end{array}\right.\]

\emph{(Note: Initialize with $\vec{\alpha}_{0} = 0$.)}

%\item[2.8.b)] Optional: (LARS-OLS hybrid method) Compute the regression coefficients using least squares $$\vec{\alpha}_{\text{OLS},k} = \tilde{X}_S^\dagger \vec{y},$$ where $\tilde{X}_S$ is the matrix of selected, non-signum-aligned covariates.

\item[2.9)] Repeat Steps 1.1-1.8 until all covariates have zero correlation with the residual or until there are no covariates remaining in $S^c$.

\end{enumerate}
\noindent Output: Vector of regression coefficients $\vec{\alpha}$.\\
\bigskip

Step 2.2 finds the maximum absolute correlation $\hat{C}$ by examining $|\hat{c}_j|$ for $j=1,...,r$. In the original LARS algorithm, the covariates corresponding to the maximum absolute correlation are multiplied by the signs of their correlations with the residual to provide sign-aligned versions of the covariates that are added to the active set. To adapt this step for complex LARS, consider the complex signum function \cite{}.
\begin{equation}
\text{sgn}_{\mathbbm{C}}(z) = \frac{z}{|z|},
\label{eq:sgn}
\end{equation}
which is the complex extension of the real-valued sign function $ \text{sgn}(x) = x/|x|$. The $\text{sgn}_{\mathbbm{C}}(\cdot)$ function returns a complex number that lies on the unit circle. Note that if $z$ is strictly real, $\text{sgn}_{\mathbbm{C}}(z)$ agrees with the real-valued sign function $\text{sgn}(z)$. The following proposition shows that multiplying a covariate by the signum of its correlation with the residual yields a positive correlation with the residual.

\begin{theorem}
Given covariate $\vec{x}_j$ and current residual $(\vec{y}-\vec{\hat{\mu}}_S)$, the inner product $\left<s_j\vec{x}_j,\vec{y}-\vec{\hat{\mu}}_S\right>$, where $s_j = \text{sgn}_\mathbbm{C}\left(\left<\vec{x}_j,\vec{y}-\vec{\hat{\mu}}_S\right>\right)$ is a positive, real value.\\
\emph{Proof:}
\begin{align*}
\left<s_j\vec{x}_j,\vec{y}-\vec{\hat{\mu}}_S\right> &= \left<\frac{\left<\vec{x}_j,\vec{y}-\vec{\hat{\mu}}_S\right>}{\left|\left<\vec{x}_j,\vec{y}-\vec{\hat{\mu}}_S\right>\right|}\vec{x}_j,\vec{y}-\vec{\hat{\mu}}_S\right>\\
&= \overline{\frac{\left<\vec{x}_j,\vec{y}-\vec{\hat{\mu}}_S\right>}{\left|\left<\vec{x}_j,\vec{y}-\vec{\hat{\mu}}_S\right>\right|}} \left<\vec{x}_j,\vec{y}-\vec{\hat{\mu}}_S\right>\\
&= \left|\left<\vec{x}_j,\vec{y}-\vec{\hat{\mu}}_S\right>\right| > 0.
\end{align*} \QED
\label{thrm:pos_corr}
\end{theorem}

Proposition \ref{thrm:pos_corr} shows that the signum of the complex correlation provides a method of aligning a covariate so that it is positively correlated with the residual. Using the signum-aligned covariates in the active set, the next step of LARS is the construction of an equiangular search direction. The next proposition provides the new direction of travel.

\begin{theorem}
Let an equi-inner product condition for covariates in the active set be given by
\begin{equation}
X_S^H \vec{u}_S = L_S \vec{\mathbbm{1}}. \label{eqn:equiIP}
\end{equation}
The resulting inner product value $L_S$, active set covariate weights $\vec{w}_S$, and the equi-inner product direction $\vec{u}_S$ are, respectively,
\begin{align}
L_S &= \left(\vec{\mathbbm{1}}^T\left(X_S^H X_S\right)^{-1}\vec{\mathbbm{1}}\right)^{-1/2},\label{eqn:Ls}\\
\vec{w}_S &= \left(X_S^H X_S\right)^{-1}L_S\vec{\mathbbm{1}},\\
\vec{u}_S &= X_S \vec{w}_S.
\end{align}
\emph{Proof:}
Let $X_S \in \mathbbm{C}^{n\times q}$ be a nonempty active set where n is the covariates dimension and $q$ is the number of covariates. Since $X_S$ (by construction), we have that $\text{rank}(X_S) = q$. Therefore, the associated $q\times q$ Gram matrix $X_S^H X_S$ has \cite{Strang}
$$ \text{rank}(X_S^H X_S) = \text{rank}(X_S) = q.$$
The matrix $X_S^H X_S$ is therefore invertible.  Since we construct $\vec{u}_S$ to reside in the column space of $X_S$, then the projection of $\vec{u}_S$ onto the column space of $X_S$ using the projection operator $X_S(X_S^H X_S)^{-1}X_S^H$ returns $\vec{u}_S$, providing the identity
\begin{equation}
X_S(X_S^H X_S)^{-1}X_S^H \vec{u}_S = \vec{u}_S.
\label{eq:proof2}
\end{equation}
Taking the inner product of both sides with $\vec{u}_S$ and invoking the constraint $\vec{u}_S^H \vec{u}_S = 1$, yields
$$ \vec{u}_S^H X_S(X_S^H X_S)^{-1} X_S^H \vec{u}_S = 1.$$
Inserting the equi-inner product condition \eqref{eqn:equiIP} and solving leads to
$$L_S = \left(\vec{\mathbbm{1}}^T\left(X_S^H X_S\right)^{-1}\vec{\mathbbm{1}}\right)^{-1/2}.$$
Multiplying the equi-inner product condition \eqref{eqn:equiIP} on the left by $X_S(X_S^H X_S)^{-1}$ and invoking the identity \eqref{eq:proof2}
provides the travel direction
$$\vec{u}_S = X_S(X_S^H X_S)^{-1}L_S\vec{\mathbbm{1}}.$$ \QED
\label{thrm:inner_prod}
\end{theorem}

Note that the change in inner product results in minimal alteration of the calculations for the travel direction. The third proposition of this section gives the equation for length of travel along $\vec{u}_S$.

\begin{theorem}
If the active set $S$ and its complement $S^c$ are nonempty, then traveling along the equi-inner product direction $\vec{u}_S$ by the length
\begin{equation}
\hat{\gamma} = \min_{j \in S^c}\!^+ \quad \frac{\left(\Re\left\{\left<g_j,\hat{c}_j\right>\right\}-\hat{C}L_S\right) \pm \sqrt{\left(\Re\left\{\left<g_j,\hat{c}_j\right>\right\}-\hat{C}L_S\right)^2 - \left(|g_j|^2 - L_S^2\right)\left(|\hat{c}_j|^2 - \hat{C}^2\right)}}{|g_j|^2 - L_S^2},\label{eqn:gammaHat}
\end{equation}
causes a new covariate to enter the active set.\\
\emph{Proof:}
After a step $\gamma \vec{u}_S$ in the estimate, the absolute current correlation between each covariate $\vec{x}_j$ and the residual becomes,
\[
\left\lvert \left<\vec{x}_j, \vec{y} - (\vec{\hat{\mu}}_{S,k-1} + \gamma \vec{u}_S\right>\right\rvert = \left\lvert \hat{c}_j - \gamma g_j   \right\rvert,
\]
where $g_j = \left<\vec{x}_j,\vec{u}_S \right>$.  For a member of the active set $S$, the current absolute correlation becomes $\left \lvert \hat{C} - \gamma L_S\right\rvert$, since $\hat{c}_j = \hat{C}$, and $g_j = L_S$, for each $j \in S$.  Equating the absolute current correlations provides the condition
\begin{equation}
\left\lvert \hat{c}_j - \gamma g_j   \right\rvert  = \left\lvert \hat{C} - \gamma L_S \right\rvert. \label{eqn:travelDirectionCond}
\end{equation}
When condition \eqref{eqn:travelDirectionCond} holds, a new covariate achieves an equal absolute correlation and enters the active set $S$.  

Note that since $X^H X $ is positive-definite, $\left(X^H X\right)^{-1} $ is positive definite.  Further, note that $L_S$ given in \eqref{eqn:Ls} contains a quadratic form of a positive-definite matrix, and $L_S$ itself turns out to be a strictly positive, real number.  Since $\hat{C} >0$ and $L_S > 0$, the condition \eqref{eqn:travelDirectionCond} leads to the quadratic equation
\[
\left(|g_j|^2 - L_S^2\right)\gamma^2 - 2\left(\Re \left\{\left< g_j,\hat{c}_j\right>\right\} - \hat{C}L_S \right)\gamma + \left(|\hat{c}_j|^2 - \hat{C}^2\right) = 0. 
\]
The solution of this equation is given in \eqref{eqn:gammaHat} with the additional restriction that we seek $\hat{\gamma}$ to be the minimum value with $\gamma > 0$ such that condition \eqref{eqn:travelDirectionCond} holds.\QED
\label{thrm:distance}
\end{theorem}

Propositions \ref{thrm:pos_corr} through \ref{thrm:distance} provide the necessary tools for adapting LARS for complex covariates and complex data.  If the covariates exist in complex conjugate pairs, as often occurs for real-valued data, note that LARS will add both covariates in a conjugate pair to the active set during the same iteration, since the current absolute correlations are equal when $y$ and $\vec{\hat{\mu}}_S$ are real, i.e., $\left\lvert \left<\vec{x}_j,\vec{y}-\vec{\hat{\mu}}_S\right> \right\rvert = \left\lvert \left<\overline{\vec{x}}_j,\vec{y}-\vec{\hat{\mu}}_S\right> \right\rvert$.

%%%%%%%%%%%%%%%%%%%%%%%%%%%%%%%%%%%%%%%%%%%%%%%%%%%%%%%%%%%%%%%%%%%%%%%%%%%%%%%%

\section{Least Angle Regression for Dynamic Mode Decomposition}\label{sec:LARS_DMD}

DMD often provides complex DMD modes and DMD eigenvalues.  Section III provides the necessary modification of the LARS algorithm for complex quantities, and this section applies Complex LARS to DMD mode selection.  To apply Complex LARS to the problem of DMD mode selection, we seek a mapping between the DMD modes and the covariates in Complex LARS so that covariate selection equates to mode selection.  

Observe that Complex LARS takes a zero-mean data vector $\vec{y}$ as input.  Let $\text{vec}(\cdot)$ denote the operation of vectorizing a two-dimensional array into a column vector by stacking the columns of the array.  Vectorization of the data matrix $\Psi_0$ given in \eqref{eq:opt} and subtraction of the mean provides a zero-mean data vector 
\begin{equation}
\vec{y} = \text{vec}(\Psi_0) -  \text{mean}\left(\text{vec}(\Psi_0)\right),\label{eq:zeroMeanData}
\end{equation}
containing the complete temporal evolution of the measurements.  Similarly, to capture the temporal evolution of the DMD modes in vector form, consider the Vandermonde matrix $\Xi$ in \eqref{eq:vand}, which captures the temporal evolution of the eigenvalue coefficients of the DMD modes.  Define the $j$th row of $\Xi$ to be $\vec{\xi}_j \in \mathbb{C}^{1\times N}$ such that
\[
\left[
\begin{array}{c}
\vec{\xi}_1\\
\vec{\xi}_2\\
\vdots\\
\vec{\xi}_r
\end{array}
\right]
=
\left[
\begin{array}{cccc}
\lambda_1^{0}&\lambda_1^{1}&\dots&\lambda_1^{N-1}\\
\lambda_2^{0}&\lambda_2^{1}&\dots&\lambda_2^{N-1}\\
\vdots&\vdots&\ddots&\vdots \\
\lambda_r^{0}&\lambda_r^{1}&\dots&\lambda_r^{N-1}\\
\end{array}
\right].
\]
Using the Vandermonde row vectors, each covariate $\vec{x}_j$ for $j = 1,\dots,r$, can be constructed by
$$\vec{x}_j = \frac{\text{vec}\left(\vec{\phi}_j \vec{\xi}_j\right) -  \text{mean}\left(\text{vec}\left(\vec{\phi}_j \vec{\xi}_j\right)\right)}{\sqrt{\text{var} \left(\text{vec}\left(\vec{\phi}_j \vec{\xi}_j\right)\right)}},$$
where the variance $\text{var}(\vec{q}) = 1/n\sum_{j=1}^n|q_j-\text{mean}(\vec{q})|^2$ for an $n$-dimensional vector $\vec{q}$. Each resulting $\vec{x}_j$ is zero-mean, unit-variance, and lies in $\mathbb{C}^{mN\times1}$, where $m$ is the length of each snapshot and $N$ is the number of snapshots. The covariates $\vec{x}_j$ for $j=1,\dots,r,$ consist of the time resolved DMD modes but are not scaled by the DMD mode amplitudes.  Appropriate scaling of the DMD mode amplitudes is the reduced-order modeling problem. Algorithm 3 presents LARS4DMD, the reduced-order modeling algorithm proposed by this paper.

\begin{center}
\textbf{Algorithm 3 (LARS4DMD)}\\
\end{center}
\noindent Inputs: Data matrices $\Psi_0$ and $\Psi_1$.
\begin{enumerate}
\item[3.1)] Calculate the DMD modes $\vec{\phi}_j$ and the DMD eigenvalues $\lambda_j$ for $j=1,\dots,r$, from $\Psi_0$ and $\Psi_1$ using the SVD of $\Psi_0$, the eigendecomposition of \eqref{eq:Fdmd}, and equation \eqref{eq:DMDmode}.\\
\item[3.2)] Form the data vector $\vec{y}$
$$\vec{y} = \text{vec}(\Psi_0) - \text{mean}(\text{vec}(\Psi_0)).$$
\item[3.3)] Form zero-mean, unit-variance covariates
$$\vec{x}_j = \frac{\text{vec}\left(\vec{\phi}_j \vec{\xi}_j\right) -  \text{mean}\left(\text{vec}\left(\vec{\phi}_j \vec{\xi}_j\right)\right)}{\sqrt{\text{var} \left(\text{vec}\left(\vec{\phi}_j \vec{\xi}_j\right)\right)}},$$
for $j=1,\dots,r$.
\item[3.4)] Call Complex LARS, forming the LARS4DMD estimate
\[
\vec{\mu}_{\text{LD},k} = \vec{\hat{\mu}}_{S,k-1} + \frac{\hat{C}}{L_S}\vec{u}_S,
\]
within Step 2.6.  Within Step 2.7, also calculate the LARS4DMD regression coefficients
\[
\alpha^\text{LD}_{k,j} = \left\{ \begin{array}{ll} \alpha_{k-1} + \frac{\hat{C}}{L_S}s_j w_{S,j} & \text{if} j\in S,\\
0 & \text{if} j\notin S,
\end{array}\right.
\]
\end{enumerate}
\noindent Output: Vector of regression coefficients $\alpha^\text{LD}_k$ for each iteration $k$.\\
\bigskip

Once the zero-mean, unit-variance covariates $\vec{x}_j$ and the zero-mean data $\vec{y}$ are formed in Steps 3.1 through 3.3, the user can perform Complex LARS to obtain regression coefficients.  However, the standard regression coefficients from LARS assume that the algorithm is allowed to run to completion (cf. Steps 1.9  Algorithm 1 and Step 2.9 of Algorithm 2).  Generating a model using fewer modes than the total number available can be thought of as pausing the Complex LARS algorithm and temporarily discarding the remaining modes that are not in the active set (i.e., emptying $S^c$).  The LARS algorithm (Algorithm 1) provides a separate calculation for the travel distance along the equiangular direction when $S^c$ is empty, given by $\hat{\gamma} = \hat{C}/L_S$.  The authors of \cite{LARS} note that this choice results in regression coefficients that correspond with the least-squares solution assuming a model based on the covariates in the active set only.  Step 3.4 in Algorithm 3 implements this choice for travel distance when calculating the LARS4DMD estimate $\vec{\mu}_{\text{LD},k}$ and the LARS4DMD regression coefficients $\vec{\alpha}^\text{LD}_k$ at each iteration $k$.  We note that this choice of travel distance $\hat{\gamma}$ corresponds to variant of LARS known as the LARS Ordinary Least Squares (OLS) hybrid method in \cite{LARS} that performs re-fitting of the regression coefficients using least squares on the selected covariates.

 Each $j$th regression coefficient $\alpha^\text{LD}_{k,j}$ corresponds to the $j$th covariate $\vec{x}_j$, which is a manipulated version of the corresponding $j$th DMD mode $\vec{\phi}_j$ and Vandermonde row $\vec{\xi}_j$.  To generate a reconstruction of original dataset, calculate
 \begin{equation}
\text{vec}\left(\Psi^\text{LD,rec}_0\right) = X \vec{\alpha}^\text{LD}_k + \text{mean}\left(\text{vec}(\Psi_0)\right),
 \end{equation}
 where the addition of of the scalar value $\text{mean}(\text{vec}(\Psi_0))$ occurs elementwise to add back the data mean that was previously subtracted when generating $y$ in \eqref{eq:zeroMeanData}.  Reshaping the $\text{vec}\left(\Psi^\text{LD,rec}_0\right)$ vector provides the reconstruction of $\Psi_0$.  The next section performs a numerical comparison of the performance loss in the reconstuction of $\Psi_0$ by the LARS4DMD and DMDSP algorithms.

\section{Numerical experiment: Poiseuille flow} \label{sec:results}

This section compares the performance and mode selection of the DMDSP and LARS4DMD algorithms on a Poiseuille flow test case.  Pressure-driven flow in a channel between two parallel walls (of infinite depth perpendicular to the flow) is known as plane Poiseuille flow \cite{Deen}.    The Poiseuille flow solution can be obtained from the two-dimensional, linearized Navier-Stokes equations.  These equations are solved numerically at Reynolds Number $\text{Re} = 10,000$ using a pseudo-spectral scheme in \cite{DMDSP}. The first author of \cite{DMDSP} generously provides the data from the solution on the author's website at \cite{DMDSP_site}.  Please refer to \cite{DMDSP} for details on the numerical methods used in generation of the Poiseuille flow dataset.

Using the numerically generated data for Poiseuille flow from \cite{DMDSP,DMDSP_site}, we implemented the DMDSP and LARS4DMD algorithms to obtain models of various sizes.  For DMDSP, we chose the regularization parameter $\beta$ to range logarithmically from $5\times10^{-6}$ to $160$, providing $2,500$ values for $\beta$.  Although DMDSP treats each $\beta$ value as a separate optimization problem, we examine the sequence of models from DMDSP that result for decreasing $\beta$ values to compare with sequential model construction procedure of LARS4DMD.

\begin{figure}[thpb]
\centering
\includegraphics[scale=0.8]{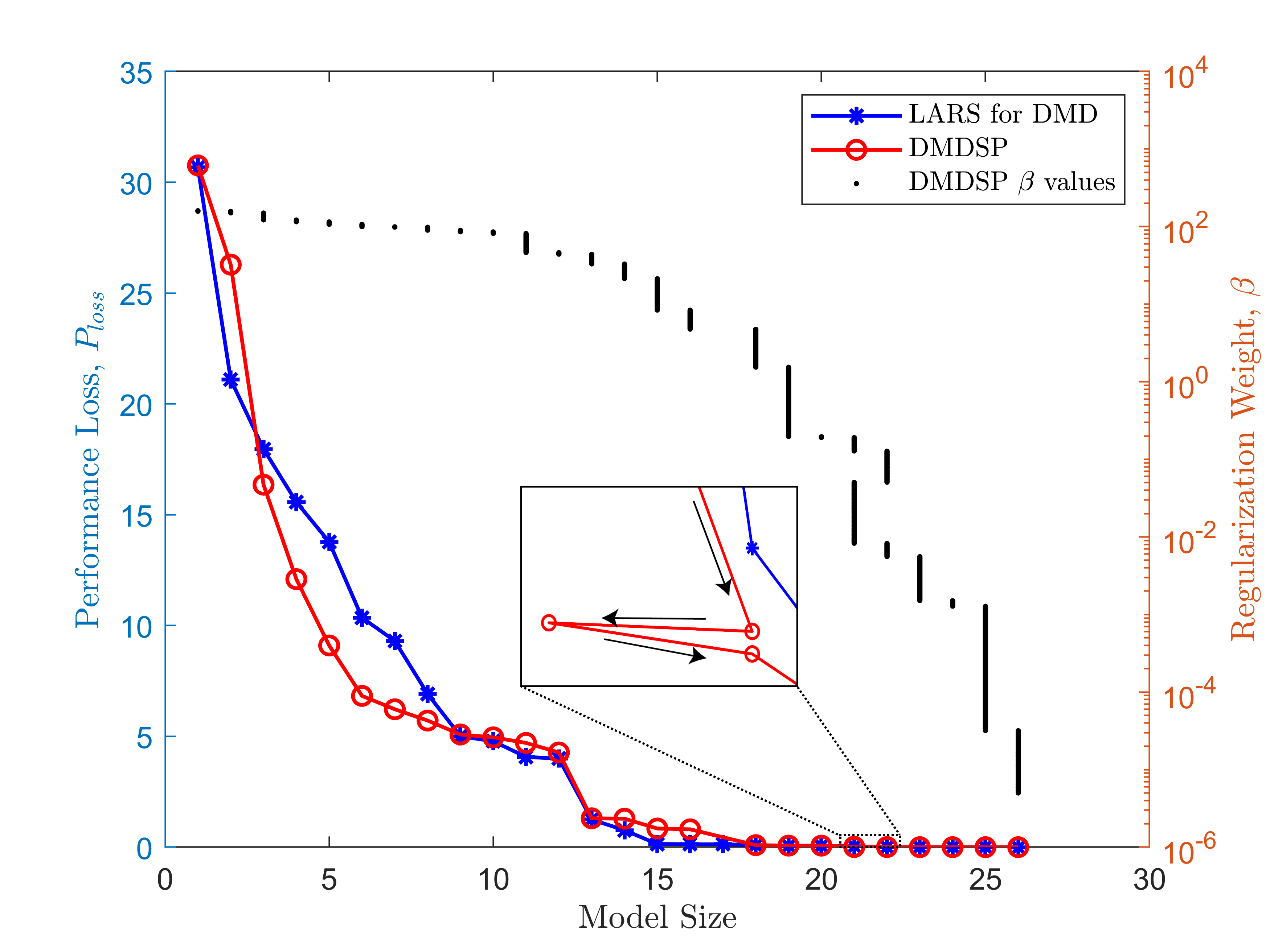}
\caption{Performance loss comparison for DMDSP and DMD with LARS for increasing model size}
\label{fig:P_loss}
\end{figure}

Figure \ref{fig:P_loss} compares the performance loss of the dataset reconstruction for these two methods.   Note that LARS4DMD and DMDSP provide the same performance loss for a one-mode model.  As $\beta$ decreases, DMDSP often produces the same model several times in a row.  After crossing a certain $\beta$ threshold, DMDSP changes model size by adding one or more modes to the model. The performance loss for DMDSP and LARS4DMD remain close for increasing model size.  For a two-mode model, LARS4DMD provides better performance.  However, for model sizes 3 through 8, DMDSP provides better performance.  LARS4DMD lies below the DMDSP performance loss for model sizes 9 through 20.   For larger model sizes, the two methods yield comparable values.  Across all model sizes, DMDSP and LARS4DMD yield similar performance loss, with DMDSP only notably outperforming LARS4DMD in the intermediate range of model sizes 3 through 8.  

Figure \ref{fig:P_loss} contains a black dot for each $\beta$ value in DMDSP, and each marker denotes a corresponding system model generated by DMDSP.  Note that there are ranges of $\beta$ values that yield models of the same size.  In total, there were only $26$ unique models (i.e., models with unique $\alpha$ vectors) out of the $2,500$ models generated by DMDSP, and there were only $24$ unique model sizes. DMDSP skips model size 17.  DMDSP generated two models containing $21$ DMD modes and two models containing $22$ DMD modes. In contrast, LARS4DMD sequentially constructed $26$ unique models by stopping the LARS procedure at each iteration to generate a new model.

\begin{figure}[H]
\centering
\subfigure[LARS4DMD\label{fig:LARS_Order}]
{
\includegraphics[width=.7\linewidth]{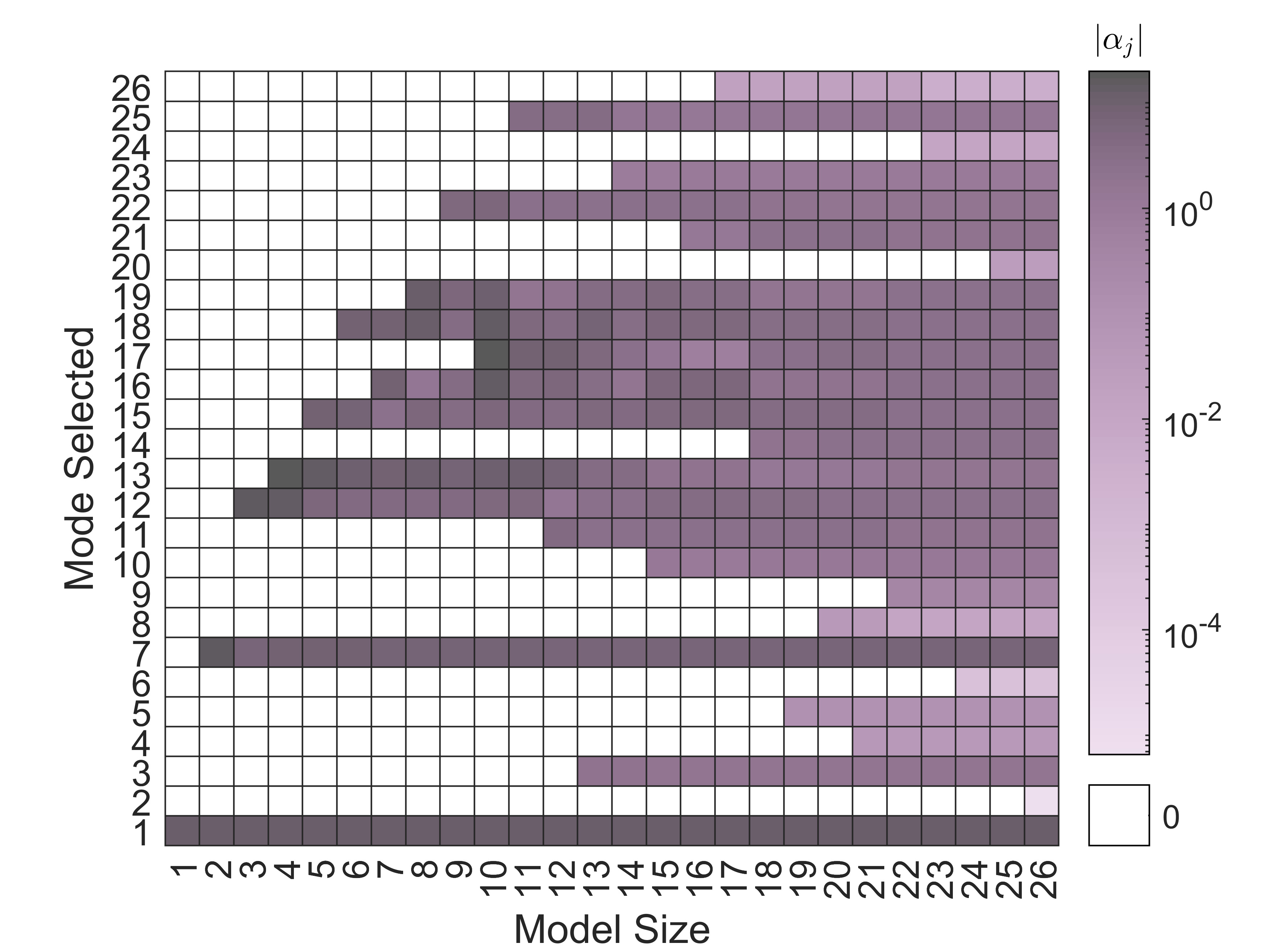}
}
\subfigure[DMDSP\label{fig:DMDSP_Order}]
{
\includegraphics[width=.7\linewidth]{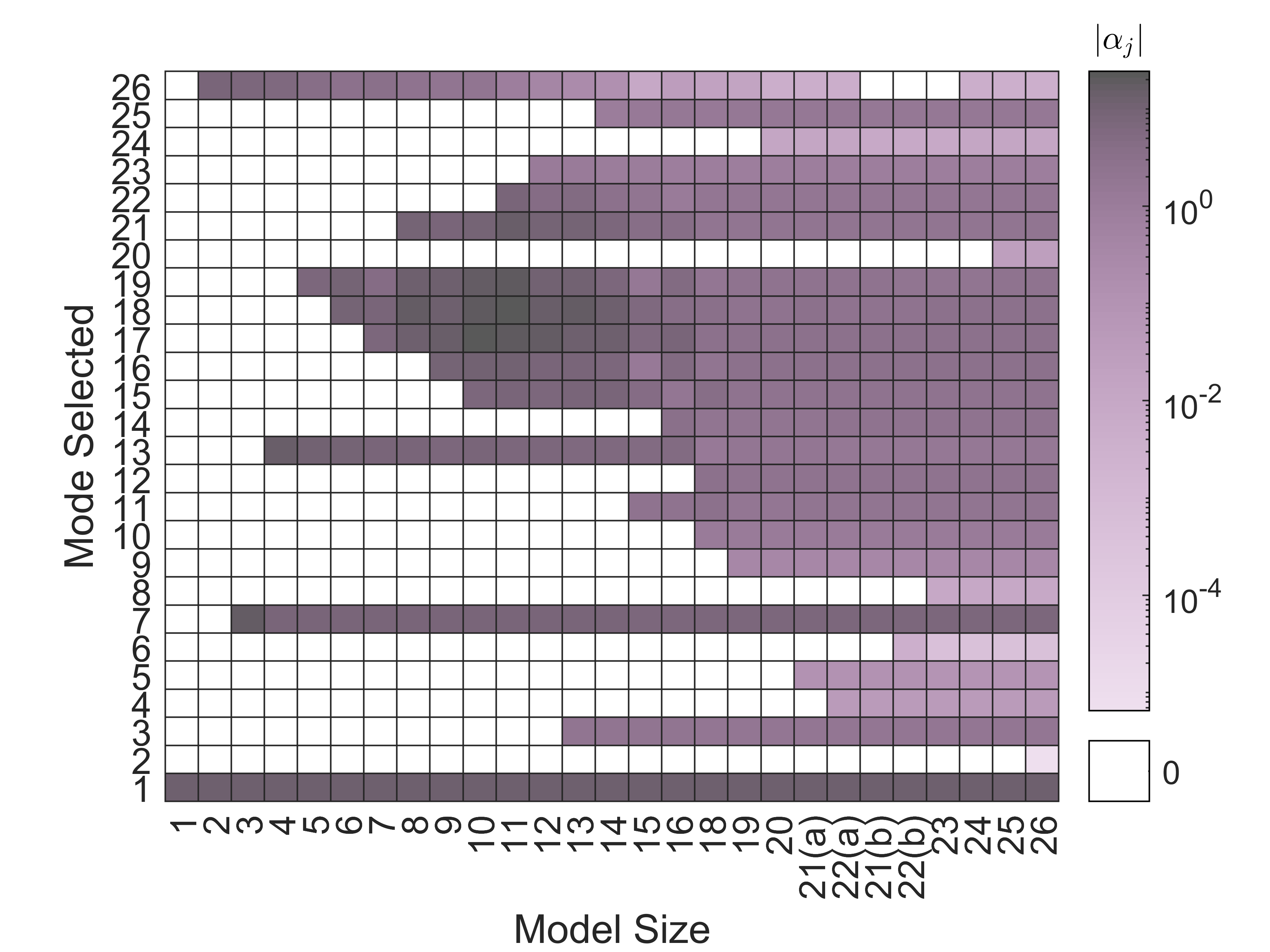}
}
\caption{Modes selected based on model size}
\end{figure}

Figures \ref{fig:LARS_Order} and \ref{fig:DMDSP_Order} illustrate the DMD modes selected in the unique models constructed by LARS4DMD and DMDSP, respectively.  The shading of each square represents the absolute value of the DMD mode amplitude $|\alpha_j|$ for the $j$th mode.  For some model sizes, LARS4DMD and DMDSP produce very similar models.  For example, the one-mode model and the three largest models match very closely between the two techniques.  Mode 1 is selected first and mode 2 is selected last by both algorithms.  Generally, both methods weight the most recently added mode the least, however exceptions occur (e.g. in Fig. \ref{fig:LARS_Order}, the addition of mode 21 at model size 16 outweighs mode 17 which was added at model size 10).  However, there are also notable differences for the models produced by these methods.  For model size 2, LARS4DMD selects mode 7, but DMDSP selects mode 26.  However, DMDSP eventually scales mode 26 to zero in model 21(b).  Note that DMDSP has the ability to remove modes from the model, because DMDSP solves a separate optimization problem for each $\beta$ value.

DMDSP produces two models each for model sizes 21 and 22. In order of increasing regularization parameter $\beta$, DMDSP increases from model size 21 to 22, then decreases back to 21 before increasing again to 22.  This sequence can be seen in inset plot of Figure \ref{fig:P_loss}. This behavior is also depicted in Figure \ref{fig:DMDSP_Order} where regularization parameter $\beta$ decreases from left to right.

%%%%%%%%%%%%%%%%%%%%%%%%%%%%%%%%%%%%%%%%%%%%%%%%%%%%%%%%%%%%%%%%%%%%%%%%%%%%%%%%

\section{Conclusion}\label{sec:conclusion}
In this paper, we adopt a useful algorithm known as Least Angle Regression (LARS) from the statistics and machine learning literature and adapt it for construction of reduced-order DMD models. Utilizing a complex inner product, we create a complex version of LARS that can be applied to select DMD modes, which are often complex.  We refer to the DMD-based reduced-order modeling algorithm as LARS4DMD.  LARS4DMD constructs a reduced-order model in a sequential manner by selecting the DMD modes that are highly correlated with the residual between the measured data and the current estimate. The results from LARS4DMD are comparable to results from Sparsity-Promoting Dynamic Mode Decomposition (DMDSP), a popular DMD-based modeling algorithm. In ongoing work, we are testing the performance of LARS4DMD on data collected experimentally for which measurement noise is present.

\bibliography{references}

\begin{thebibliography}{21}
\newcommand{\enquote}[1]{``#1''}
\providecommand{\natexlab}[1]{#1}
\providecommand{\url}[1]{\texttt{#1}}
\providecommand{\urlprefix}{URL }
\expandafter\ifx\csname urlstyle\endcsname\relax
  \providecommand{\doi}[1]{doi:\discretionary{}{}{}#1}\else
  \providecommand{\doi}{doi:\discretionary{}{}{}\begingroup
  \urlstyle{rm}\Url}\fi

\bibitem[{Schmid(2010)}]{Schmid2010}
Schmid, P., \enquote{Dynamic mode decomposition of numerical and experimental
  data,} \emph{Journal of Fluid Mechanics}, Vol. 656, No.~1, 2010, pp. 5--28.
\newblock \doi{10.1017/s0022112010001217}.

\bibitem[{Zhang et~al.(2014)Zhang, Liu, and Wang}]{ZHANG_2014}
Zhang, Q., Liu, Y., and Wang, S., \enquote{The identification of coherent
  structures using proper orthogonal decomposition and dynamic mode
  decomposition,} \emph{Journal of Fluids and Structures}, Vol.~49, 2014, pp.
  53 -- 72.
\newblock \doi{https://doi.org/10.1016/j.jfluidstructs.2014.04.002}.

\bibitem[{Brunton et~al.(2016)Brunton, Johnson, Ojemann, and
  Kutz}]{BRUNTON_2016}
Brunton, B.~W., Johnson, L.~A., Ojemann, J.~G., and Kutz, J.~N.,
  \enquote{Extracting spatial temporal coherent patterns in large-scale neural
  recordings using dynamic mode decomposition,} \emph{Journal of Neuroscience
  Methods}, Vol. 258, 2016, pp. 1 -- 15.
\newblock \doi{https://doi.org/10.1016/j.jneumeth.2015.10.010}.

\bibitem[{Surana and Banaszuk(2016)}]{koopman}
Surana, A., and Banaszuk, A., \enquote{Linear observer synthesis for nonlinear
  systems using Koopman Operator framework,} \emph{IFAC-PapersOnLine}, Vol.~49,
  No.~18, 2016, pp. 716 -- 723.
\newblock \doi{https://doi.org/10.1016/j.ifacol.2016.10.250}, 10th IFAC
  Symposium on Nonlinear Control Systems NOLCOS 2016.

\bibitem[{Bai et~al.(2019)Bai, Kaiser, Proctor, Kutz, and
  Brunton}]{Brunton_AIAA_2019}
Bai, Z., Kaiser, E., Proctor, J.~L., Kutz, J.~N., and Brunton, S.~L.,
  \enquote{Dynamic Mode Decomposition for compressive system identification,}
  \emph{AIAA Journal}, 2019, pp. 1--14.
\newblock \doi{10.2514/1.J057870}.

\bibitem[{Liu et~al.(2019)Liu, Tan, and Cao}]{LIU_2019}
Liu, M., Tan, L., and Cao, S., \enquote{Dynamic mode decomposition of
  cavitating flow around ALE 15 hydrofoil,} \emph{Renewable Energy}, Vol. 139,
  2019, pp. 214 -- 227.
\newblock \doi{https://doi.org/10.1016/j.renene.2019.02.055}.

\bibitem[{Pan et~al.(2017)Pan, Wang, Wang, and Sun}]{Pan_2017}
Pan, C., Wang, J., Wang, J., and Sun, M., \enquote{Dynamics of an unsteady
  stagnation vortical flow via dynamic mode decomposition analysis,}
  \emph{Experiments in Fluids}, Vol.~58, No.~3, 2017, p.~21.
\newblock \doi{10.1007/s00348-017-2306-1}.

\bibitem[{Tu et~al.(2014)Tu, Rowley, Luchtenburg, Brunton, and Kutz}]{Tu2014}
Tu, J.~H., Rowley, C.~W., Luchtenburg, D.~M., Brunton, S.~L., and Kutz, J.~N.,
  \enquote{On Dynamic Mode Decomposition: Theory and applications.}
  \emph{Journal of Computational Dynamics}, Vol.~1, No.~2, 2014, pp. 391 --
  421.

\bibitem[{Jovanovic et~al.(2014)Jovanovic, Schmid, and Nichols}]{DMDSP}
Jovanovic, M.~R., Schmid, P.~J., and Nichols, J.~W.,
  \enquote{Sparsity-promoting Dynamic Mode Decomposition,} \emph{Physics of
  Fluids}, Vol.~26, No.~2, 2014, p. p. 024103.
\newblock \doi{10.1063/1.4863670}.

\bibitem[{{Le Ngo} et~al.(2017){Le Ngo}, {See}, and {Phan}}]{Emotion}
{Le Ngo}, A.~C., {See}, J., and {Phan}, R.~C., \enquote{Sparsity in Dynamics of
  Spontaneous Subtle Emotions: Analysis and application,} \emph{IEEE
  Transactions on Affective Computing}, Vol.~8, No.~3, 2017, pp. 396--411.
\newblock \doi{10.1109/TAFFC.2016.2523996}.

\bibitem[{Murata et~al.(2018)Murata, Aihara, Tokuda, Iwamitsu, Mizoguchi, Akai,
  and Okada}]{Phonon}
Murata, S., Aihara, S., Tokuda, S., Iwamitsu, K., Mizoguchi, K., Akai, I., and
  Okada, M., \enquote{Analysis of coherent phonon signals by Sparsity-Promoting
  Dynamic Mode Decomposition,} \emph{Journal of the Physical Society of Japan},
  Vol.~87, No.~5, 2018, p. 054003.
\newblock \doi{10.7566/JPSJ.87.054003}.

\bibitem[{{Annoni} et~al.(2016){Annoni}, {Seiler}, and
  {Jovanović}}]{DMDSP_inputs}
{Annoni}, J., {Seiler}, P., and {Jovanović}, M.~R.,
  \enquote{Sparsity-Promoting Dynamic Mode Decomposition for systems with
  inputs,} \emph{2016 IEEE 55th Conference on Decision and Control (CDC)},
  2016, pp. 6506--6511.
\newblock \doi{10.1109/CDC.2016.7799270}.

\bibitem[{Dawson et~al.(2016)Dawson, Hemati, Williams, and Rowley}]{Dawson2016}
Dawson, S. T.~M., Hemati, M.~S., Williams, M.~O., and Rowley, C.~W.,
  \enquote{Characterizing and correcting for the effect of sensor noise in the
  Dynamic Mode Decomposition,} \emph{Experiments in Fluids}, Vol.~57, No.~3,
  2016, p.~42.
\newblock \doi{10.1007/s00348-016-2127-7}.

\bibitem[{Gomez et~al.(2019)Gomez, Lagor, Kirk, Lind, Jones, and Paley}]{Gomez}
Gomez, D.~F., Lagor, F., Kirk, P.~B., Lind, A., Jones, A.~R., and Paley, D.~A.,
  \emph{Unsteady DMD-based flow field estimation from embedded pressure sensors
  in an actuated airfoil}, 2019.
\newblock \doi{10.2514/6.2019-0346}.

\bibitem[{Efron et~al.(2004)Efron, Hastie, Johnstone, and Tibshirani}]{LARS}
Efron, B., Hastie, T., Johnstone, I., and Tibshirani, R., \enquote{Least Angle
  Regression,} \emph{Annals of Statistics}, Vol.~32, No.~2, 2004, pp. 407--451.
\newblock \doi{10.1214/009053604000000067}.

\bibitem[{DMD(2019)}]{DMDSP_site}
\enquote{DMDSP-Sparsity-Promoting Dynamic Mode Decomposition,} , 2019.
\newblock \urlprefix\url{www.umn.edu/~mihailo/software/dmdsp/}, accessed:
  2019-04-25.

\bibitem[{Kutz et~al.(2017)Kutz, Brunton, Brunton, and Proctor}]{DMD_Book}
Kutz, J.~N., Brunton, S.~L., Brunton, B.~W., and Proctor, J.~L., \emph{Dynamic
  mode decomposition: data-driven modeling of complex systems}, Society for
  Industrial and Applied Mathematics, 2017.

\bibitem[{Lebedev and Cloud(2003)}]{inner_prod}
Lebedev, L.~P., and Cloud, M.~J., \emph{The calculus of variations and
  functional analysis: With optimal control and applications in mechanics},
  World Scientific Publishing Co Pte Ltd, Singapore, SINGAPORE, 2003.

\bibitem[{D'Angelo(2002)}]{herm_prod}
D'Angelo, J.~P., \emph{Inequalities from complex analysis}, American
  Mathematical Society, Washington, USA, 2002.

\bibitem[{Strang(2006)}]{Strang}
Strang, G., \emph{Linear algebra and its applications}, 4\textsuperscript{th}
  ed., Thomson Brooks/Cole, Belmont, CA USA, 2006.

\bibitem[{Deen(1998)}]{Deen}
Deen, W.~M., \emph{Analysis of transport phenomena}, Oxford University Press,
  New York, NY USA, 1998.

\end{thebibliography}
\end{document}